\documentclass[sigconf]{acmart}

\AtBeginDocument{%
  }

\usepackage{multirow}
\usepackage{listings}
\usepackage{caption}

\begin{document}

\title{Integrating Summarization and Retrieval for Enhanced Personalization via Large Language Models}




\author{Christopher Richardson}
\authornote{Both authors contributed equally to this research.}
\affiliation{%
  \institution{Georgia Institute of Technology}
  \city{Atlanta}
  \state{GA}
  \country{USA}
}
\email{crichardson332@gmail.com}
\orcid{1234-5678-9012}

\author{Yao Zhang}
\authornotemark[1]
\affiliation{%
  \institution{Amazon Alexa AI}
  \city{Austin}
  \state{TX}
  \country{USA}
}
\email{yaozhanq@amazon.com}

\author{Kellen Gillespie}
\affiliation{%
  \institution{Amazon Alexa AI}
  \city{Seattle}
  \state{WA}
  \country{USA}
}
\email{kelleng@amazon.com}

\author{Sudipta Kar}
\affiliation{%
  \institution{Amazon Alexa AI}
  \city{Seattle}
  \state{WA}
  \country{USA}
}
\email{sudipkar@amazon.com}

\author{Arshdeep Singh}
\affiliation{%
  \institution{Amazon Alexa AI}
  \city{Seattle}
  \state{WA}
  \country{USA}
}
\email{adpsingh@amazon.com}

\author{Zeynab Raeesy}
\affiliation{%
  \institution{Amazon Alexa AI}
  \city{Seattle}
  \state{WA}
  \country{USA}
}
\email{raeesyzr@amazon.com}

\author{Omar Zia Khan}
\affiliation{%
  \institution{Amazon Alexa AI}
  \city{Seattle}
  \state{WA}
  \country{USA}
}
\email{ozkhan@amazon.com}

\author{Abhinav Sethy}
\affiliation{%
  \institution{Amazon Alexa AI}
  \city{Seattle}
  \state{WA}
  \country{USA}
}
\email{sethya@amazon.com}

\renewcommand{\shortauthors}{Richardson and Zhang et al.}

\begin{abstract}
Personalization, the ability to tailor a system to individual users, is an essential factor in user experience with natural language processing (NLP) systems. With the emergence of Large Language Models (LLMs), a key question is how to leverage these models to better personalize user experiences. To personalize a language model's output, a straightforward approach is to incorporate past user data into the language model prompt, but this approach can result in lengthy inputs exceeding limitations on input length and incurring latency and cost issues. Existing approaches tackle such challenges by selectively extracting relevant user data (i.e. selective retrieval) to construct a prompt for downstream tasks. However, retrieval-based methods are limited by potential information loss, lack of more profound user understanding, and cold-start challenges. To overcome these limitations, we propose a novel summary-augmented approach by extending retrieval-augmented personalization with task-aware user summaries generated by LLMs. The summaries can be generated and stored offline, enabling real-world systems with runtime constraints like voice assistants to leverage the power of LLMs. Experiments show our method with 75\% less of retrieved user data is on-par or outperforms retrieval augmentation on most tasks in the LaMP personalization benchmark. We demonstrate that offline summarization via LLMs and runtime retrieval enables better performance for personalization on a range of tasks under practical constraints.
\end{abstract}

\ccsdesc[500]{Computing methodologies~Natural language generation}
\ccsdesc[500]{Information systems~Personalization}
\ccsdesc[300]{Computing methodologies~Information extraction}

\keywords{LLM, Personalization, Summarization, NLP, Chatbot, Voice Assistant, Conversational AI}


\maketitle

\section{Introduction}
As virtual assistants and other natural language processing (NLP) systems become increasingly integrated into our daily lives, personalization has become an essential factor in user experience. Tailoring virtual assistant interactions and NLP model outputs to individual users' preferences, styles, needs, and contexts is essential in improving the performance of these systems to make them more natural and conversational.

Traditional personalization methods, such as collaborative filtering \cite{schafer_collaborative_2007}, deep neural networks \cite{covington_deep_2016}, deep interest network \cite{zhou_deep_2018} and their variations \cite{qi_search-based_2020, chen_end--end_2021}, have enhanced user experiences in recommendation systems. These methods leverage historical user behavior data to make personalized recommendations, offering a practical and effective solution for various domains. Despite their success, these methods still struggle with the cold-start problem, where new users lack sufficient behavior history, leading to sub-optimal recommendations. The cold-start problem highlights the need for alternative approaches.


Large Language Models (LLMs) represent a promising avenue for advancing personalization techniques. LLMs have demonstrated remarkable capabilities in understanding context and generating coherent text \cite{brown_language_2020}. By incorporating knowledge about users, LLMs can potentially enhance personalization by capturing subtle user preferences, but how to capture the full spectrum of user preferences in a personalized manner remains a challenge. To personalize a language model output, a straightforward approach is to incorporate user data into the language model prompt. However, incorporating a comprehensive view of customer preferences with long-term historical user data into the prompt may exceed the input length limitations of language models and result in considerable increases in inference cost. Further, language models tend to degrade with lengthy contexts \cite{chen2023demonstrations}. 
To address these concerns, a personalized retrieval augmentation framework was proposed \cite{salemi2023lamp}. This framework selectively extracts relevant user data to construct prompts for downstream language models. Recent work has also shown promise in combining retrieval approaches with LLMs to improve performance in recommender systems \cite{chen2023palr,xu2022rethinking,li2023gpt4rec}, as well as general NLP tasks \cite{dudy2022personalization,flek2020returning,qian2021pchatbot,rafieian2022ai,wu2021personalized}.
However, retrieval-based methods have constraints in potential information loss, lack the ability to comprehend user data on a more profound level, and may suffer from the cold-start problem.

\begin{figure}
    \centering
    \includegraphics[width=0.4\textwidth]{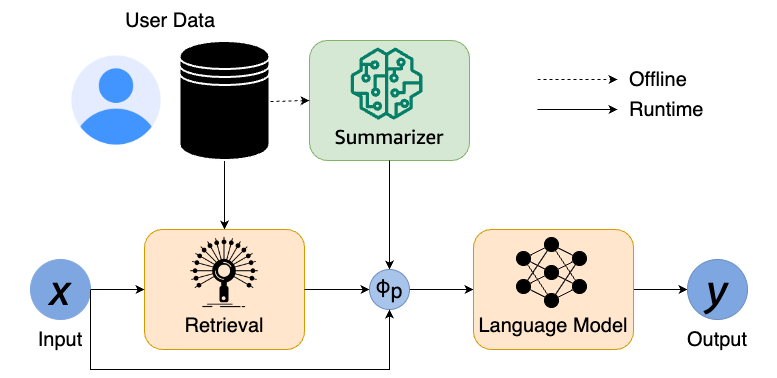}
    \caption{Personalization is achieved by combining runtime-retrieved samples with an offline-generated user summary. Given a textual
input $x$ that describes a task in natural language, the goal is to generate a personalized output $y$ for users. The retrieval model identifies the most relevant items from user data, and the retrieved items along with the offline user summary and $x$ form the basis for creating a prompt. This prompt is constructed using a prompt construction function $\phi_p$.
}
    \label{fig:main}
\end{figure}

Our research aims to address the aforementioned limitations of both traditional personalization methods and retrieval-based methods with LLMs by proposing the hybrid approach shown in Figure \ref{fig:main}. By integrating retrieval techniques with LLM-generated summaries of user data, we intend to create a more robust personalized system. 
To prevent information loss, the user summary offers contextual information at a higher level of abstraction for the downstream task.
To understand user data on a more profound level,  the summary generation is aware of the task and incorporates this information in the prompt for summary generation. For example, for a personalized paraphrase text generation task, the summary model is instructed by a prompt to pay attention to the user writing style in addition to the semantic content.
Also, this hybrid model could overcome the cold-start problem and provide personalized outputs even in data-sparse scenarios by providing user summaries for new users based on available user data from other applications or user's self description.
The summaries in our approach can be generated offline and stored, ensuring negligible increased runtime latency and enabling systems with runtime constraints to leverage the power of LLMs into real-work online applications, such as voice assistant scenarios.

We demonstrate our method of integrating summarization and retrieval on a publicly available Language Model Personalization (LaMP) benchmark \cite{salemi2023lamp}, including both text classification and generation tasks across a variety of domains. 
Experiments show our method achieves comparable or better performance compared to retrieval augmentation on most tasks. 
With our method, the retrieval component can use 75\% less of retrieved user data without sacrificing performance on five out of six tasks, and achieves superior performance on two tasks.

In summary, our main contributions are as follows. First, we propose augmenting traditional retrieval-based personalization methods with LLMs' summarization of user data to address the limitations of existing methods: potential information loss, the inability to understand user data at a high level, and the cold-start challenge. Our method enables powerful LLMs to provide comprehensive information about users with no additional runtime latency. Further, we implemented our proposed approach and conducted experiments on a language model personalization benchmark dataset LaMP with 6 public tasks. With promise shown in our experiment results, we envision a personalized system that better caters to individual user preferences especially for new users by integrating summarization via LLMs and retrieval.

\section{Methodology}
\subsection{Problem Formulation}
The problem formulation follows LaMP \cite{salemi2023lamp}: given a textual input $x$ that describes a task in natural language, we want to generate a personalized output $y$ for user $u$. The goal is thus to learn the distribution $p(y|x, u)$. 

\subsubsection{\textbf{Baseline}}
Our baseline is a retrieval-augmented method that follows a retrieve-then-model paradigm \cite{salemi2023lamp}. For retrieval, a manually defined query generation function $\phi_q(x)$ is first used to help extract salient information from $x$ as inputs to the retrieval model $\mathcal{R}(\phi_q(x),P_u,k)$. The retrieval model returns the top-$k$ relevant items from the user profile $P_u$ using the retrieval query $\phi_q(x_i)$, and the returned items are used to construct a prompt for a downstream model using a prompt construction model $\phi_p$. The input to the downstream language model is as follows.
\begin{align}
    \overline{x}_i = \phi_p(x_i, \mathcal{R}(\phi_q(x_i),P_u,k))
\end{align}
The downstream language model is fine-tuned on the dataset $\{\overline{x}_i, y_i\}$.

\subsection{Integrating Summarization and Retrieval}
Runtime constraints can limit the number of user data retrieved by $\mathcal{R}$ that can be utilized. In this work, we consider scenarios where there are both latency considerations as well as input length limits for the downstream model. To improve performance without adding runtime latency, we introduce a summary of the user data, $s_u$, to augment the retrieved data: 
\begin{align}
    \overline{x}_i = \phi_p(x_i, \mathcal{R}(\phi_q(x_i),P_u,k), s_u)
\end{align}
Our approach involves using LLMs to summarize salient information from $u$ as it relates to optimizing $p(y|x, u)$. We use instruction-tuned models to generate an abstractive summary of user data:
\begin{align}
    s_u = \text{LLM}(P_u)
\end{align}

An overview of our method is shown in Figure \ref{fig:main}. The summaries can be generated offline and stored along with the user data itself. At runtime, the retrieval algorithm retrieves the top-$k$ profile entries and concatenates them with the task input and the summary to create the full context for the downstream language model, which is fine-tuned using the standard language modeling loss against the output $y$.

\section{Experiments}

\subsection{\textbf{Datasets and Evaluation Metrics}}
LaMP is a public benchmark dataset for training and evaluating methods for personalization with language models \cite{salemi2023lamp}. It consists of seven personalization NLP tasks, including three classification tasks and four text generation tasks. Data for each task include input text, reference completion text as ground truth output, and a user profile consisting of an array of items with textual data. A brief description of each task and their evaluation metrics are shown in Table \ref{tab:lamp}. We have excluded task LaMP-6 from our study as it relies on private data to which we do not have access. The LaMP benchmark organizes the data in both user-based separation and time-based separation. In our study, we  utilize the user-based separation approach to address cold-start issues for new users.

\begin{table}[]
    \caption{Description of LaMP tasks and data.}
    \label{tab:lamp}
    \centering
    \begin{tabular}{cp{2.6in}}
        \toprule
        \multirow{3}{*}{LaMP-1} & \textbf{Task}: Citation Identification (binary choice) \\
        & \textbf{User Profile}: Scholarly article titles and abstracts \\
        & \textbf{Metric(s)}: Accuracy \\ \midrule
        \multirow{3}{*}{LaMP-2} & \textbf{Task}: News Categorization (classification) \\
        & \textbf{User Profile}: Categorized articles published  \\
        & \textbf{Metric(s)}: Accuracy and F1 \\ \midrule
        \multirow{3}{*}{LaMP-3} & \textbf{Task}: Product Rating (classification) \\
        & \textbf{User Profile}: Product reviews and scores \\
        & \textbf{Metric(s)}: Mean Absolute Error (MAE) and Root Mean Square Error (RMSE) \\ \midrule
        \multirow{3}{*}{LaMP-4} & \textbf{Task}: News Headline Generation (text generation) \\
        & \textbf{Profile}: News articles and their headlines \\
        & \textbf{Metric(s)}: ROUGE-1 and ROUGE-L \\ \midrule
        \multirow{3}{*}{LaMP-5} & \textbf{Task}: Scholarly Title Generation (text generation) \\
        & \textbf{Profile}: Scholarly article titles and abstracts \\
        & \textbf{Metric(s)}: ROUGE-1 and ROUGE-L \\ \midrule
        \multirow{3}{*}{LaMP-7} & \textbf{Task}: Tweet Paraphrasing (text generation) \\
        & \textbf{Profile}: Tweets \\
        & \textbf{Metric(s)}: ROUGE-1 and ROUGE-L \\
        \bottomrule  
    \end{tabular}
\end{table}

\subsection{Experimental Setup}
Following LaMP, we used FlanT5-base \cite{chung2022scaling} as our downstream model. This model demonstrated satisfactory runtime performance in our experiments (approximately 125 milliseconds per sample when we included as many user data as possible within the 512-token limit of the input length) and, as shown in the LaMP experiments, it achieved superior performance to that achieved in zero-shot experiments with FlanT5-XXL and ChatGPT \cite{salemi2023lamp}. For all experiments, we used the same settings reported in LaMP: a learning rate of $5 \times 10^{-5}$, weight decay of $10^{-4}$, warmup ratio of 0.05, and a beam size of 4, and we trained for 10 epochs for text classification (tasks 1-3) and 20 epochs for text generation (tasks 4, 5, and 7). 

In our experiments, we utilized the BM25 retrieval algorithm \cite{robertson1995okapi} due to its speed and performance. 
We found neural methods like Contriever \cite{asai2022task} to be too slow for voice assistant scenarios, which induced approximately 10-30 seconds of latency per sample while not significantly outperforming BM25 on many LaMP tasks.

We experimented with two instruction-tuned models for generating summaries. The first is Vicuna \cite{zheng2023judging}, a 13-billion parameter model distilled from LLaMA \cite{touvron2302llama}. The second model is ChatGPT using OpenAI's API with the \lstinline{gpt-3.5-turbo-16k} model. Vicuna has a context length of 2048 tokens, while ChatGPT's limit is 16,384. Prompts used for generating summaries are shown in Table \ref{tab:prompts}. For tasks 2 and 3, to achieve good performance given the simplicity of the tasks, we constrained the summarization model to output according to a strict template shown in the Table \ref{tab:prompts}, therefore ChatGPT summaries were not included for these tasks.

We compare our methods to the retrieval-only baselines, using $k$ values of 0, 1, and 4 for the baselines. We observed $k=4$ to be the limit for some tasks given the context length of FlanT5 (512 tokens). Thus, we had to reduce the number of retrieved samples to fit the summaries into the input of FlanT5 without truncating and chose $k=1$ (for direct comparison with the baseline), as well as $k=0$ to investigate the impact of summaries alone (no retrieval). We report means of three repeated runs of each experiment for comparison with statistical significance.

\begin{table*}[]
    \caption{Prompts used for summarization. Additional tokens were used for Vicuna summaries to match the expected prompt format for that model but the content was the same for Vicuna and ChatGPT.}
    \label{tab:prompts}
    \centering
    \begin{tabular}{cp{6in}} 
        \toprule
        Task & Prompt \\ \midrule
        LaMP-1 & Write a summary, in English, of the research interests and topics of a researcher who has published the following papers. Only generate the summary, no other text. \\ 
        LaMP-2 & Look at the following past articles this journalist has written and determine the most popular category they write in. Answer in the following form: most popular category: <category>\\
        LaMP-3 & Based on this user's past reviews, what are the most common scores they give for positive and negative reviews? Answer in the following form: most common positive score: <most common positive score>, most common negative score: <most common negative score>\\
        LaMP-4 & Given this author's previous articles, try to describe a template for their headlines. I want to be able to accurately predict the headline gives one of their articles. Be specific about their style and wording, don't tell me anything generic.\\
        LaMP-5 & Given this author's previous publications, try to describe a template for their titles. I want to be able to accurately predict the title of one of the papers from the abstract. Only generate the template description, nothing else.\\
        LaMP-7 & Given this person's previous tweets, try to describe a template for their tweets. I want to take a generic sentence and rephrase it to sound like one of their tweets, with the same style/punctuation/capitalization/wording/tone/etc. as them. Only give me the template description, nothing else. \\ 
        \bottomrule  
    \end{tabular}
\end{table*}

\section{Results and Discussion}

Table \ref{tab:results} shows the results of our methods for both summary models alongside the baselines across various NLP tasks. Our experiments prove our summary-augmented method with $k=1$ is on-par or outperforms the retrieval-only baseline with $k=4$ on most tasks (reducing the amount of retrieved user data by 75\%). As the bold results indicate the best results among compared experiments for each task, our methods (Vicuna Summ. and GPT-3.5 Summ.) outperform baselines consistently on tasks 1 and 2 at a statistical significance level with $p-value<0.05$, and we achieve comparable performance on tasks 3, 5, and 7 with no statistically significant difference.

On LaMP-1 task, our method GPT-3.5 Summ. outperforms the baseline with $k=4$ using only offline generated summaries ($k=0$). Furthermore, Our method GPT-3.5 Summ. yields better results compared to $k=1$ baseline on all tasks. Worth noting that ChatGPT summaries mostly outperform those provided by Vicuna, likely due to the disparity in model size.

\begin{table*}[]
    \centering
    \caption{Results for FlanT5-base model fine-tuned on LaMP benchmark tasks. Baseline: retrieval of k user data entries; Vicuna Summ.: Baseline + summary of user data generated by vicuna; GPT-3.5 Summ.: Baseline + summary of user data generated by GPT-3.5. \underline{Underline} means summary improved the corresponding baselines with the same k, and \textbf{bold} means the best results among compared experiments for each task. For all metrics, higher is better except in the case of MAE and RMSE used for LaMP-3.}
    \label{tab:results}
    \begin{tabular}{l l c c c c c c c}
    \toprule
        & & \multicolumn{3}{c}{Baseline} & \multicolumn{2}{c}{Vicuna Summ.} & \multicolumn{2}{c}{GPT-3.5 Summ.} \\ \cmidrule(lr){3-5} \cmidrule(lr){6-7} \cmidrule(lr){8-9}
        Task & Metric & $k=0$ & $k=1$ & $k=4$ & $k=0$ & $k=1$ & $k=0$ & $k=1$ \\ \midrule
        LaMP-1: Personalized       & \multirow{2}{*}{Accuracy} & \multirow{2}{*}{0.516} & \multirow{2}{*}{0.650} & \multirow{2}{*}{0.709} & \multirow{2}{*}{\underline{0.704}} & \multirow{2}{*}{\underline{0.728}} & \multirow{2}{*}{\underline{0.738}} & \multirow{2}{*}{\textbf{\underline{0.743}}} \\ 
        Citation Identification \\ \midrule
        LaMP-2: Personalized       & Accuracy  & 0.731 & 0.782 & 0.807 & \underline{0.801} & \textbf{\underline{0.814}} & \multicolumn{2}{c}{\multirow{2}{*}{N/A}} \\ 
        News Categorization        & F1        & 0.511 & 0.573 & 0.574 & \underline{0.550} & \underline{\textbf{0.601}} & & \\ \midrule
        LaMP-3: Personalized       & MAE       & 0.311 & 0.284 & 0.280 & \underline{0.305} & \textbf{\underline{0.277}} & \multicolumn{2}{c}{\multirow{2}{*}{N/A}} \\ 
        Product Rating             & RMSE      & 0.626 & 0.595 & \textbf{0.593} & 0.632 & \underline{0.594} & & \\ \midrule
        LaMP-4: Personalized       & ROUGE-1   & 0.152 & 0.177 & \textbf{0.188} & \underline{0.157}	& 0.173 & \underline{0.170} & \underline{0.181} \\ 
        News Headline Generation   & ROUGE-L   & 0.137 & 0.162 & \textbf{0.173} & \underline{0.142} & 0.159 & \underline{0.155} & \underline{0.166} \\ \midrule
        LaMP-5: Personalized       & ROUGE-1   & 0.424 & 0.447 & \textbf{0.448} & \underline{0.426} & 0.447 & 0.424 & \textbf{\underline{0.448}} \\ 
        Scholarly Title Generation & ROUGE-L   & 0.382 & 0.408 & \textbf{0.409} & \underline{0.386} & 0.408& \underline{0.383} & \textbf{\underline{0.409}} \\ \midrule
        LaMP-7: Personalized       & ROUGE-1   & 0.510 & 0.502 & \textbf{0.513} & 0.510 & \underline{0.512} & \underline{0.510} & \underline{0.512} \\ 
        Tweet Paraphrasing         & ROUGE-L   & 0.455 & 0.448 & 0.459 & 0.455 & \underline{0.459} & \underline{0.456} & \textbf{\underline{0.460}} \\

     \bottomrule
    \end{tabular}
\end{table*}

We observed a gap between Vicuna Summ. and GPT-3.5 Summ. attributed to the differing quality of the offline summaries.. Despite studies showing Vicuna achieving up to 90\% the performance of ChatGPT \cite{vicuna2023}, our results suggest that Vicuna did not perform as well as GPT-3.5 on task LaMP-1 and 4. To assess the summary quality, we provide examples in the Appendix. \ref{examples}.

While we have shown promise in combining offline summaries with runtime retrieval for personalization, there are a few limitations to this work. For one, the data and tasks provided in the LaMP benchmark are simplistic and narrow in scope. More work is needed to assess the potential of our method on more realistic user data. Also, the benefits of summarization can be improved by fine-tuning a larger language model and end-to-end training for the tasks.









\section{Conclusion}
This paper introduces a novel method for augmenting retrieval with offline summarization for improving personalization in various NLP tasks. We implemented our method and achieved comparable or better performance on most NLP tasks in the LaMP personalization benchmark while reducing the amount of retrieved user data by 75\%. In some cases, we even achieved superior performance after removing retrieval entirely, showing an advantage for sparse data scenarios, such as the cold-start problem. Our method efficiently leverages offline summary generation and is suitable for runtime constrained applications such as voice assistants.

\begin{acks}
Thanks to Camden Parsons, Farideh Tavazoee, Xingzhi (Jacky) Guo for reviewing and providing feedbacks on our initial paper draft.
\end{acks}

\bibliographystyle{ACM-Reference-Format}
\bibliography{references}

\appendix
\section{Summary Examples}\label{examples}
We are presenting examples in which the LaMP baseline model exhibited errors, while our methods demonstrated superior performance in Example 1, and in which the LaMP baseline model and Vicuna Summ. exhibited errors while GPT-3.5 Summ. demonstrated superior performance in Example 2.


\subsection{Example 1}

\textbf{Task: LaMP-1}

\textbf{Input:}
For an author who has written the paper with the title "QuickSense: Fast and energy-efficient channel sensing for dynamic spectrum access networks", which reference is related? Just answer with [1] or [2] without explanation. [1]: "A first look at cellular network performance during crowded events" [2]: "On the complexity of scheduling in wireless networks"

\textbf{Output:} [1]

\textbf{Retrieved user data:}
On the Stability of Rational, Heterogeneous Interdomain Route Selection

A decision-theoretic approach to resource allocation in wireless multimedia networks

A general algorithm for interference alignment and cancellation in wireless networks

Enterprise social network analysis and modeling: A tale of two graphs

\textbf{Baseline Prediction:} [2]

\textbf{User Summary by Vicuna:}
(The original output is in Chinese and here is the translation:)
Network Traffic Analysis

Wireless Communication

Software-Defined Networking

Traffic Routing

Low-Latency Communication

Multimedia Networks

Network Security

Error Handling in Networks

Sensor Networks

Multi-Path Routing

IP Routing

Traffic Analysis Tools

\textbf{Vicuna Summ. Prediction:} [1]

\textbf{User Summary by GPT-3.5:}
The research interests of this researcher span several areas in computer networking and wireless communications. They have published papers on topics such as unified cellular and ad-hoc network architecture, optimization of wireless mesh networks, networking analytics, incentive-compatible routing protocols, software-defined networking, and network security. They have expertise in optimizing network throughput, designing scalable network architectures, developing game-theoretic and cryptographic techniques for wireless networks, and leveraging machine learning and data analytics for network monitoring and diagnosis. Their recent works also focus on addressing the latency and energy efficiency challenges in wireless multimedia networks and leveraging software-defined networking for next-generation cellular networks. In summary, this researcher's work revolves around developing novel algorithms, protocols, and systems that improve the performance, resilience, and security of networked systems.

\textbf{GPT-3.5 Summ. Prediction:} [1]

\subsection{Example 2}

\textbf{Task: LaMP-1}

\textbf{Input:}
For an author who has written the paper with the title "Visual-audio integration for user authentication system of partner robots", which reference is related? Just answer with [1] or [2] without explanation. [1]: "Pronunciation Modeling for Improved Spelling Correction" [2]: "Path following algorithm for highly redundant manipulators"

\textbf{Output:} [1]

\textbf{Retrieved user data:}
Advancing Matrix Completion by Modeling Extra Structures beyond Low-Rankness

Improved Asymmetric Locality Sensitive Hashing (ALSH) for Maximum Inner Product Search (MIPS)

Real-Time Implementation of Improved State-Space MPC for Air Supply in a Coke Furnace

On Practical Algorithms for Entropy Estimation and the Improved Sample Complexity of Compressed Counting

\textbf{Baseline Prediction:} [2]

\textbf{User Summary by Vicuna:}
1. Nystrom Method for Approximating the GMM Kernel.

2. Very sparse random projections.

3. Very sparse stable random projections for dimension reduction in lalpha (0 \&lt;alpha\&lt;=2) norm.

4. A Comparison Study of Nonlinear Kernels.

5. One sketch for all: Theory and Application of Conditional Random Sampling.

6. Collaborative Multi-objective Ranking.

7. Accurate Estimators for Improving Minwise Hashing and b-Bit Minwise Hashing.

8. A new space for comparing graphs.

9. Stability and Risk Bounds of Iterative Hard Thresholding.

10. Hashing Algorithms for Large-Scale Learning.

11. Variational Flow Graphical Model.

\textbf{Vicuna Summ. Prediction:} [2]

\textbf{User Summary by GPT-3.5:}
The researcher's primary research interests revolve around the development of practical algorithms for entropy estimation, compressed counting, and matrix completion. They explore various techniques for improving the efficiency and accuracy of these methods. Additionally, the researcher focuses on advancing hashing algorithms, particularly in the context of maximum inner product search (MIPS) and locality-sensitive hashing (LSH). They also investigate sketch-based sampling techniques, especially for sparse data, and propose novel approaches for estimating entropy of data streams. Furthermore, the researcher has contributions in optimization methods, classification, and dimensionality reduction using stable random projections. Their work demonstrates a strong emphasis on efficient and scalable solutions for a wide range of machine learning and data analysis tasks.

\textbf{GPT-3.5 Summ. Prediction:} [1]

\end{document}